\documentclass[runningheads]{llncs}

\usepackage{graphicx}
\usepackage{comment}

\usepackage{algorithm}
\usepackage[noend]{algpseudocode}
\usepackage{subfigure}

\begin{document}
\title{Enriching Visual with Verbal Explanations for Relational Concepts -- Combining LIME with Aleph}
\titlerunning{Enriching Visual with Verbal Explanations for Relational Concepts}
%
\author{Johannes Rabold \and
Hannah Deininger \and
Michael Siebers \and
Ute Schmid}
\authorrunning{J. Rabold et al.}
%
\institute{Cognitive Systems, University of Bamberg, Germany}
\maketitle              
\begin{abstract}
With the increasing number of deep learning applications, there is a growing demand for explanations. Visual explanations provide information about which parts of an image are relevant for a classifier's decision. However, highlighting of image parts (e.g., an eye) cannot capture the relevance of a specific feature value for a class (e.g., that the eye is wide open). Furthermore, highlighting cannot convey whether the classification depends on the mere presence of parts or on a specific spatial relation between them. Consequently, we present an approach that is capable of explaining a classifier's decision in terms of logic rules obtained by the Inductive Logic Programming system Aleph. The examples and the background knowledge needed for Aleph are based on the explanation generation method LIME. We demonstrate our approach with images of a blocksworld domain. First, we show that our approach is capable of identifying a single relation as important explanatory construct. Afterwards, we present the more complex relational concept of towers. Finally, we show how the generated relational rules can be explicitly related with the input image, resulting in richer explanations.

\end{abstract}
\section{Introduction}



Explainable Artificial Intelligence (XAI) mostly refers to visual highlighting of information which is relevant for the classification decision of a given instance \cite{gunning2017explainable,samek2017explainable}. In general, the mode of an explanation can be visual, but also verbal or example-based \cite{Miller19}.
Visual explanations have been introduced to make black-box classifiers such as (deep) neural networks more transparent \cite{gunning2017explainable,samek2017explainable,ribeiro2016should}. In the context of white-box machine learning approaches, such as decision trees or Inductive Logic Programming (ILP) \cite{muggleton2018ultra}, it is argued that these models are already transparent and interpretable by humans \cite{muggleton2018ultra}. In the context of ILP it has been shown that a local verbal explanation can easily be generated from symbolic rules with a template-based approach \cite{SiebersS19}.

For image classification tasks, it is rather obvious that visual explanations are helpful for technical as well as for domain experts: Information about what pixels or patches of pixels most strongly contribute to a class decision can help to detect model errors which might have been caused by non-representative sampling. Highlighting can also support domain experts to assess the validity of a learned model \cite{ribeiro2016should}.
In general, an explanation can be characterized as useful, if it meets the principles of cooperative conversations \cite{Miller19}. These pragmatic aspects of communication are described in the Gricean maximes \cite{grice1975logic} which encompass the following four categories: (1) quality -- explanations should be based on truth or empirical evidence; (2) quantity -- be as informative as required; (3) relation -- explanations should communicate only relevant information; (4) manner -- avoidance of obscurity and ambiguity.
We argue that visual explanations can in general not avoid obscurity and ambiguity since they cannot or only partially capture the following kinds of information:
{\small
\begin{itemize}
\item \textbf{Feature values:} Visual highlighting can explain that a specific aspect of an entity is informative for a specific class -- e.g., that an emotion is expressed near the eye. However, the relevant information is whether the eye is wide open or the lid is tightened \cite{siebers2016characterizing}.
\item \textbf{Negation:} While approaches like LRP \cite{samek2017explainable} allow to visualize which pixels have a negative contribution to the classification, it is not generally possible to inform that the absence of a feature or object is relevant. E.g., it might be relevant to explain that a person is not classified as a terrorist because he or she does not hold a weapon (but a flower).
\item \textbf{Relations:} If two parts of an image are highlighted, it is not possible to discriminate whether the conjunction (e.g., there is a green block and a blue block) or a more specific relation (e.g., the green block is on the blue block) is relevant.
\end{itemize}}
ILP approaches \cite{Muggleton94} can capture all three kinds of information because the models are expressed as first-order Horn clauses. Relational concepts such as \textit{grandparent(X,Y)} \cite{muggleton2015meta} or mutagenicity of chemical structures \cite{srinivasan1996theories} can be induced. Furthermore, classes involving relations, such as the Michalski Train Domain \cite{muggleton2015meta}, can be learned. Here, the decision whether a train is east- or west-bound depends on relational information of arbitrary complexity, e.g., that a waggon with six wheels needs to be followed by a waggon with an open top.

Recently, there have been proposed several deep learning approaches to tackle relational concepts, such as the differentiable neural computer \cite{graves2016hybrid}, Rel\-NNs \cite{kazemi2018relnn}, or RelNet \cite{bansal2017relnet}. In contrast to ILP, these approaches depend on very large sets of training examples and the resulting models are black-box.
A helpful explanation interface should be able to take into account visual/image-based domains as well as abstract/graph-based domains. The model agnostic approach of LIME \cite{ribeiro2016should} provides linear explanations based on sets of super-pixels or words. This is not sufficient when more expressive relational explanations are necessary.
Current focus of our work is to provide relational explanations for black-box, end-to-end classifiers for image-based domains. 
We believe that for image-based domains, a combination of visual and verbal explanations is most informative with respect to the Gricean maximes. Psychological experiments also give evidence that humans strongly profit from a combination of visual and verbal explanations \cite{mayer1994whom}.

\begin{figure}[t]
	\centering
	\subfigure[A house, because three windows left of each other.]{\label{fig:janssen_fabrik}\includegraphics[width=57mm]{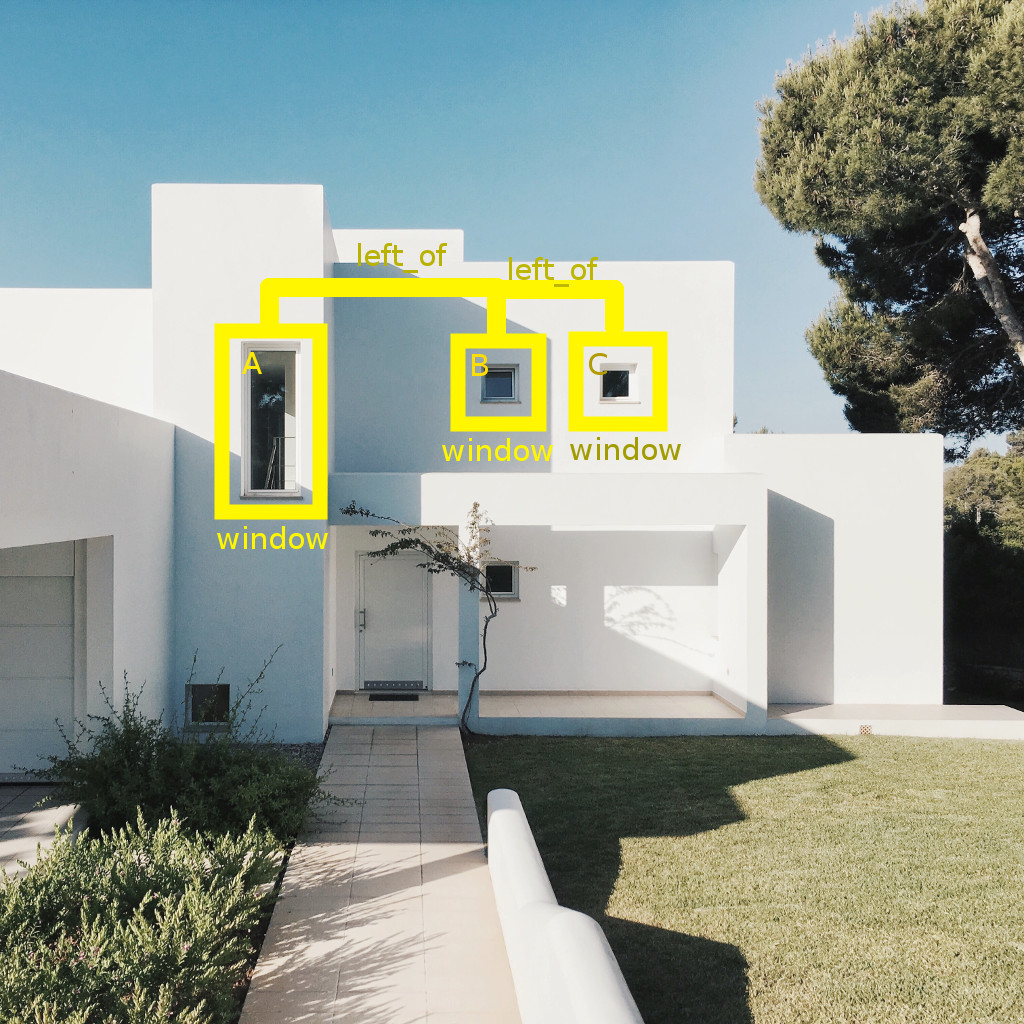}}
	\subfigure[A tower, because three windows on top of each other.]{\label{fig:roter_turm}\includegraphics[width=57mm]{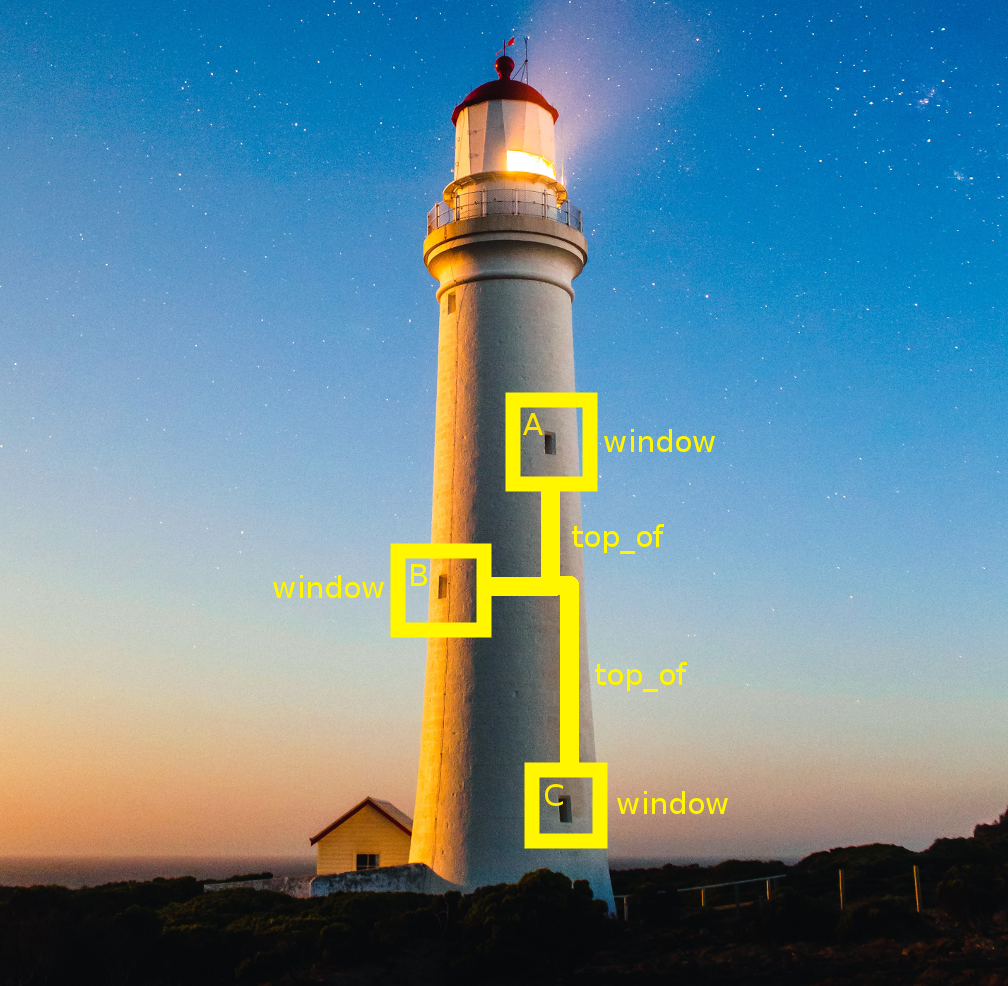}}
	\caption{\label{fig:visualverbal} Combining visual and symbolic explanations for house in contrast to tower. {\scriptsize Photo of house by Pixasquare, photo of lighthouse by Joshua Hibbert, both on Unsplash}}
\end{figure}

In a previous study \cite{RaboldSS18}, we could show that relational symbolic explanations (Prolog rules) can be generated by combining the ILP approach Aleph \cite{srinivasan2001aleph} with LIME \cite{ribeiro2016should}. However, simple visual concepts have been pre-defined and used as input to Aleph and not extracted automatically. In the following, we present an extension of \cite{RaboldSS18} covering end-to-end image classification with a convolutional neural network (CNN) \cite{krizhevsky2012imagenet}, partitioning images into sub-structures, as well as automatic extraction of visual attributes and spatial relations. Local symbolic explanations are learned with Aleph, providing logical descriptions of original and perturbed images. Finally, local symbolic explanations are related to visual highlighting of informative parts of the image to provide a combined visual-symbolic explanation. The symbolic explanation can be transformed in a verbal one with a template-based approach as demonstrated in \cite{SiebersS19}. An illustrative example is given in Figure~\ref{fig:visualverbal}. Here the concept of house is explained by the fact that three windows are next to each other. This information is given by identifying three relevant parts of the image, naming them ($A$, $B$, $C$), labeling them as windows (which might be done by another automatic image classification or by the user) and stating the spatial relation between the objects using the \texttt{left\_of} relation. This example also demonstrates an important aspect of symbolic explanations: Which attributes and relations are useful to explain why some object belongs to some class depends on the contrasting class \cite{gentner1994structural}.

In the next section, we introduce the core concepts for our approach. Then we present a significantly extended version of the LIME-Aleph algorithm \cite{RaboldSS18}. We demonstrate the approach on images of a blocksworld domain. In a first experiment we show that LIME-Aleph is capable of identifying a single relation (\texttt{left\_of(block1, block2)}) as relevant for the learned concept. In a second experiment, we demonstrate that more complex relational concepts such as \texttt{tower} can be explained. Finally we show how the fusion of visual and symbolic explanations might be realized.

\section{Explaining Relational Concepts with LIME-Aleph}

\subsection{Core Concepts}

\subsubsection{The ILP System Aleph.}

To find symbolic explanations for relational concepts we use Aleph \cite{srinivasan2001aleph}. Aleph infers a logic theory $T$ given a set of positive ($E^+$) and negative ($E^-$) examples. An example is represented by the target predicate (e.g. \texttt{stack(e1).} or \texttt{not stack(e2).}) together with additional predicates (e.g. \texttt{contains(b1, e1).}) as background knowledge (BK). Predicates in BK are used to build the preconditions for the target rules.
Aleph is based on specific-to-general refinement search. It finds rules covering as many positive examples as possible, avoiding covering negative ones. Search is guided by modes which impose a language bias. The general algorithm is \cite{srinivasan2001aleph}:
{\small
\begin{enumerate}
	\item As long as positive  exist, select one. Otherwise halt.
	\item Construct the most-specific clause that entails the selected example and is within the language constraints
	\item Find a more-general clause which is a subset of the current literals in the clause.
	\item Remove  covered by the current clause.
	\item Repeat from step 1.
\end{enumerate}}
An example of a rule from $T$ in Prolog is {\small \texttt{stack(Stack) :- contains(Block1, Stack), contains(Block2, Stack),} \texttt{on(Block2, Block1).}} \\denoting that a \texttt{stack} is defined by one block on top of another.

\subsubsection{LIME's Identification of Informative Super-Pixels.}
LIME (\textbf{L}ocal \textbf{I}nterpre\-table \textbf{M}odel-Agnostic \textbf{E}xplanations) is an approach to explain the decision result of any learned model \cite{ribeiro2016should}. Explanations state the parts of an instance that are positively or negatively correlated to the class. It works by creating a simpler, local surrogate model around the instance to be explained. In case of an image, the explanation is a set of connected pixel patches called \textit{super-pixels}.

Let $x$ be an image and $x'$ be the binary vector that states whether super-pixels $x'_i \in x'$ are switched on or off (see below). LIME finds a sparse linear model $g(x')$ that locally approximates the unknown decision function $f(x)$ represented by a black-box classifier. It effectively finds the coefficients $\vec{w}$ for the super-pixel representations being variables in a simplified linear model. This is done by generating a pool of perturbed examples $z'$ by taking the original super-pixel representation $x'$ and randomly selecting elements in a uniformly distributed fashion. That way, images $z$ are obtained with some super-pixels still original and some altered according to a transform function $h$ effectively removing the information they contained (Switching them off). Each sample $z'$ (The binary vector indicating if super-pixels are switched off in this sample) is stored in a sample pool $\cal{Z}$ along with the classifier result $f(z)$ and a distance measure $\pi_x(z)$ that expresses the distance of the perturbed example $z$ to the original image $x$. For images this can be the Mean Squared Error. The distance is needed for the linear model to be locally faithful to the original function $f(x)$ and thus has to be minimized. The ``un-faithfulness'' of the model $g$ to the black-box model $f$ with respect to the distance measure $\pi_x(z)$ is expressed with the following formula \cite{ribeiro2016should}:

\begin{center}
	$\mathcal{L}(f, g, \pi_x) = \sum_{z, z' \in \mathcal{Z}} \pi_x(z) (f(z) - g(z'))^2$.
\end{center}

\noindent The goal is to find the coefficients $\vec{w}$ for $g$ that minimize this un-faithfulness $\mathcal{L}$. The coefficients ultimately translate back to weights for the super-pixels. LIME uses K-Lasso to find the weights~\cite{efron2004hastie}.

The original LIME uses the algorithm Quick Shift \cite{vedaldi2008quick} to find super-pixel. It imposes an irregular pixel mask over the input image that segments it in terms of pixel similarity. The segmentation is performed in a 5D space consisting of the image space and the color space. Quick Shift is only one of several segmentation algorithms that are available for LIME. They all share the attribute of imposing an irregular mask over an image. In many domains, this irregularity is not wanted. For the domain used in this paper it is preferable to use a segmentation algorithm that divides an image into a regular grid with square cells.


\begin{figure}[t]
	\centering
	\subfigure[All positions in the image, where a block has to be located in order to be \textbf{l}eft, \textbf{r}ight, \textbf{t}op, or \textbf{b}ottom of block a.]{\label{fig:rel_diagram_a}\includegraphics[width=30mm]{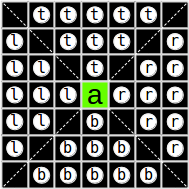}}
	\quad\quad\quad
	\subfigure[ for the relations \texttt{on} and \texttt{under}, illustrated with \texttt{on(b,a)}, \texttt{on(c,d)}, resp.~\texttt{under(a,b)} and \texttt{under(d,c)}.]{\label{fig:rel_diagram_b}\includegraphics[width=30mm]{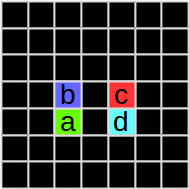}}
	\caption{\label{fig:relation_diagrams} Diagrams to show the different concepts of the relations used.}    
\end{figure}

\subsection{Extraction of Image Parts and Their Relations}\label{relex}
Based on image segmentation into a grid of super-pixels $i$ with domain-specific cell size, a set of attributes $A_i$ for cells and spatial relations between cells can be automatically extracted. Attributes $A_i$ are taken from a pool of attributes $\cal{A}$. An example for an attribute in $\cal{A}$ is the mean color of $i$ in the RGB color space. To find a human-comprehensible name, the nearest color according to the Euclidean distance in a pool of commonly known color names is assigned. Other extractable attributes are the size or the general location in the image. The coordinates of the center point of $i$ are stored for spatial reasoning. Extracted attributes are converted into predicates for BK. The attribute that a given super-pixel \texttt{SP} is blue is represented as \texttt{has\_color(SP, blue).}

Spatial relations can be defined between pairs of super-pixels. To restrict the number of pairs, we need a pre-selection $S$ of super-pixels that might be relevant for the concept. LIME's $\vec{w}$ describe the magnitude of relevance for either the true classification (positive weight) or the counter-class (negative weight). By introducing a user-defined constant $k$, we restrict how many super-pixels the selection $S$ should contain, taking the $k$ super-pixels with the highest values in $\vec{w}$. Spatial relations $r: S \times S$ are drawn from a pre-defined pool $\cal{R}$. For this work, we use the relations \texttt{left\_of, right\_of, top\_of, bottom\_of} as well as relations that represent an immediate adjacency in the regular grid mentioned earlier, namely \texttt{on} and \texttt{under}. Relations are defined with respect to the center coordinates of the super-pixels in $S$. Figure~\ref{fig:relation_diagrams} sketches the underlying semantics of these relations. It is possible to include additional relations as long as they are automatically extractable and their inverses are defined in the image space. In domains with super-pixels that differ in size, a \texttt{larger} relation between super-pixels could be defined. Also, a \texttt{not\_equal} relation can be considered.

\subsection{Learning Rules for Relational Concepts via Aleph}

To generate symbolic relational explanations for visual domains, we combine LIME's super-pixel weighting with Aleph's theory generation. The input into LIME-Aleph is an image $x$ and a model $f$ returning class probability estimations for $x$. Currently our approach is only applicable for explaining one class in contrast to all other classes, effectively re-framing the original classification as a concept learning problem. The output of LIME-Aleph is a theory $T$ of logic rules describing the relations between the super-pixels that lead to the class decision.


\begin{algorithm}[t]
	\caption{\label{alg:lime-aleph} Explanation Generation with LIME-Aleph.}
	\begin{algorithmic}[1]
		\State \textbf{Require:} Instance $x \in X$
		\State \textbf{Require:} Classifier $f$,  Selection size $k$, Threshold $\theta$
		\State \textbf{Require:} Attribute pool $\cal{A}$, Relation pool $\cal{R}$
		\State ~~~$S \leftarrow LIME(f, x, k)$ \algorithmiccomment{Selection of $k$ most important super-pixels.}
		\State ~~~$A \leftarrow$ extract\_attribute\_values$(S, \cal{A})$ \algorithmiccomment{Find all attribute values $A_i$ for all $i \in S$.}
		\State ~~~$R \leftarrow$ extract\_relations$(S, \cal{R})$ \algorithmiccomment{Find all relations $r: S \times S$ between all $i \in S$.}
		\State ~~~$E^+ \leftarrow \{\langle A, R \rangle\}$
		\State ~~~$E^- \leftarrow \{\}$
		\State ~~~\textbf{for each} $r(i,j) \in R$ \textbf{do}
		\State ~~~~~~$z \leftarrow$ flip\_in\_image$(x,i,j)$ \algorithmiccomment{Flip the super-pixels in the image space.}
		\State ~~~~~~$r' \leftarrow r(j,i)$ \algorithmiccomment{Obtain new predicate for the BK by flipping parameters.}
		\State ~~~~~~$R' \leftarrow R \setminus \{r\} \cup \{r'\}$ \algorithmiccomment{All relations in the BK; also the altered one.}
		\State ~~~~~~$R' \leftarrow calculate\_side\_effects(R', r')$ \algorithmiccomment{Re-calculate relations that are affected by the flipped relation.}
		\State ~~~~~~$c' \leftarrow f(z)$ \algorithmiccomment{Obtain new estimator for the perturbed image.}
		\State ~~~~~~\textbf{if} $c' \geq \theta$ \textbf{do} \algorithmiccomment{If estimator reaches threshold, add new positive example.}
		\State ~~~~~~~~~$E^+ \leftarrow E^+ \cup \{\langle A, R' \rangle\}$
		\State ~~~~~~\textbf{else} \algorithmiccomment{Else, add negative example.}
		\State ~~~~~~~~~$E^- \leftarrow E^- \cup \{\langle A, R' \rangle\}$
		\State ~~~\textbf{end for}
		\State ~~~ $T \leftarrow$ Aleph$(E^+, E^-)$ \algorithmiccomment{Obtain theory $T$ with Aleph.}
		\State \textbf{return} $T$
	\end{algorithmic}
\end{algorithm}

LIME's explanation relies on that linear surrogate model which contains the set of super-pixels with the highest positive weights for the true class. When dealing with the question which relations contribute most to the classification, identifying the most informative super-pixels has to be replaced by identifying the most informative \emph{pairs} of super-pixels. Instead of turning super-pixels on and off, LIME-Aleph inverts extracted relations between super-pixels and observes the effects on the classification. Algorithm~\ref{alg:lime-aleph} shows our approach. Given the selection $S$ of super-pixels together with the extracted attributes, our approach first finds all relations $R \subseteq \cal{R}$ that hold between them. For every relation $r(i,j) \in R$, a new perturbed example $z$ from the image space is created by flipping the super-pixels $i$ and $j$ in the image space. To generate a new example for Aleph, the resulting perturbed image is first put through the classifier $f$. If the estimator $f(z)$ exceeds a threshold $\theta$ for the class we want to explain, a new positive example is declared. Otherwise, the example is declared negative. All relations holding for the perturbed image are written in the BK characterizing this example. The initial positive example for Aleph is always generated for the unaltered constellation.

\section{Experiments and Results}

We investigate the applicability of LIME-Aleph in a blocksworld domain consisting of differently colored squares that can be placed in a regular-grid world. For a first investigation, we decided to focus on artificially generated images rather that real world domains.

\subsection{An Artificial Dataset for Relational Concepts} 
We implemented a generator to create a huge variety of positive and negative example images for different blocksworld concepts.
All generated images are of size $32 \times 32$ pixels consisting of a single-colored red background, containing constellations of colored squares of dimension $4 \times 4$ pixels. The squares are single-colored (excluding red) with color-channel values either being set to 0.0 or 0.8. The squares are placed into the image according to an $8 \times 8$ uniform grid.
Positive examples are generated by first randomly placing a reference square. Then, the other squares are placed randomly following the relation conventions shown in Figure~\ref{fig:relation_diagrams}. For the experiments we restricted $\cal{A} = $ \texttt{\{color\}} with attribute values in \texttt{\{cyan, green, blue\}} and $\cal{R} = $ \texttt{\{left\_of, right\_of, top\_of, bottom\_of, on, under\}}.

\subsection{Training a Black-Box Model}

To obtain a black-box model for image classification, we used a small convolutional neural network \cite{krizhevsky2012imagenet} which we trained from scratch with commonly known best-practice hyper-parameters. The network consists of two convolution layers with kernel size $2 \times 2$ and ReLU activations. Each layer learns 16 filters to be able to robustly recognize the colored squares. After flattening the output, the convolution layers are followed by 2 fully connected layers each with a ReLU activation. The first layer consists of 256 neurons, the second one of 128 neurons. A small amount of dropout is applied past each layer to cope with potential overfitting \cite{srivastava2014dropout}.
The network does not contain a pooling layer. That way, fewer location information is lost in aggregation during the learning process which we believe is crucial for preserving spatial relationships~(see \cite{goodfellow2016deep} p. 331).
For the experiments, we generated perfectly balanced datasets with 7.000 training-, 2000 validation- and 1000 test-images for both the concept and the counter-examples. We trained the networks for a maximum of 10 epochs with early stopping if the validation loss did not decrease after 5 epochs.

\subsection{Experiment 1: Single Relation Concept}

\begin{figure}[b]
	\centering
	\subfigure[]{\label{fig:first_ex_a_pos}\includegraphics[width=20mm]{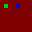}}
	\quad\quad\quad
	\subfigure[]{\label{fig:first_ex_b_pos}\includegraphics[width=20mm]{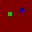}}
	\quad\quad\quad
	\subfigure[]{\label{fig:first_ex_c_neg}\includegraphics[width=20mm]{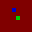}}
	\caption{\label{fig:first_example} Positive ($a$, $b$) and negative ($c$)  for the first experiment.}
\end{figure}

The concept for the first experiment can be described by the single relation that a green square is left of a blue square in an image $x$. Figure~\ref{fig:first_example} shows two positive examples ($a$, $b$) and one negative example ($c$).
After the full 10 epochs, the accuracy on the validation set reached \texttt{93.47\%}. For image \ref{fig:first_example}.a the classifier gave the estimator for the concept to be \texttt{89.83\%}. For image \ref{fig:first_example}.b the estimator was \texttt{94.18\%}. The estimator output for belonging to the concept for \ref{fig:first_example}.c was \texttt{0.28\%} showing that the network is able to discriminate the positive and negative examples.
To generate explanations for these three images, each image is separately fed into LIME. The number of kept super-pixels $k$ is set to 3. We choose this value for $k$ because we were aware that there are 2 squares in the image that are distinguishable from the background. One additional super-pixel was taken to generate a richer pool for selection $S$ containing also some background. In general, for many domains it is not that easy to estimate good values for $k$. So in most of the cases it is preferable to over-estimate the value to not lose information for the explanation.

Finally, a symbolic explanation is generated with LIME-Aleph. We describe the procedure for the positive example \ref{fig:first_example}.a. First, Algorithm~\ref{alg:lime-aleph} extracts the colors of the selected super-pixels and the relations between them. Then, all the relations get flipped one after the other to produce the example set and BK for Aleph.
The original example \ref{fig:first_example}.a is used as the seed for a set of perturbed versions of the image. Threshold $\theta$ indicates whether the perturbed example is classified positive or negative. Based on the final validation accuracy of the trained $f$ and from the estimator for the original image \ref{fig:first_example}.a, it was set to $\theta = 0.8$.
For example \ref{fig:first_example}.a 3 positive and 4 negative examples were created. From these 7 examples, Aleph induced a theory $T$ consisting of a single rule with accuracy of \texttt{100\%}:

\begin{center}
	\texttt{concept(A) :- contains(B,A), has\_color(B,green), contains(C,A), has\_color(C,blue), left\_of(B,C).}
\end{center}

\noindent The learned rule accurately resembles the construction regulation of the wanted concept; a green square has to be left of a blue square in an example A. Also, this explanation matches the input image.

For image \ref{fig:first_example}.b we used the same hyper-parameters ($k=3$, $\theta=0.8$). Again, Aleph came up with an accuracy of \texttt{100\%} and a rule structurally different, but conveying the same concept as the first rule:

\begin{center}
	\texttt{concept(A) :- contains(B,A), has\_color(B, blue), contains(C,A), has\_color(C, green), left\_of(C,B).}
\end{center}

\subsection{Experiment 2: Tower Concept}\label{complex_tower}

\begin{figure}[t]
	\centering
	\subfigure[]{\label{fig:second_ex_a_pos}\includegraphics[width=20mm]{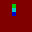}}
	\quad\quad\quad
	\subfigure[]{\label{fig:second_ex_b_neg}\includegraphics[width=20mm]{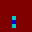}}
	\quad\quad\quad
	\subfigure[]{\label{fig:second_ex_c_neg}\includegraphics[width=20mm]{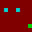}}
	\caption{\label{fig:tower_experiments} Positive ($a$) and negative ($b$, $c$)  for the tower experiment.}
\end{figure}

In the second experiment, we investigated a specific concept of towers. Positive examples consist of three differently colored blocks with a given restriction on their stacking order. An example belongs to the concept \texttt{tower}, if a blue square is present as a foundation. Directly \texttt{on} the foundation (one grid cell above) there has to be a square of either cyan or green color. Directly \texttt{on} that square has to be the remaining square (green or cyan). Figure~\ref{fig:tower_experiments} gives a positive and two negative examples.


We again trained the CNN for 10 epochs. The final validation accuracy was \texttt{98.70\%}. The original estimator for example \ref{fig:tower_experiments}.a gave $f(a) = 94.88\%$. We first set $k=3$ being the smallest selection of which we know can contain a tower. Again setting $\theta=0.8$, LIME-Aleph came up with 5 positive and 6 negative examples (accuracy \texttt{81.82\%}) and the following rule:

\begin{center}
	\texttt{concept(A) :- contains(B,A), has\_color(B,cyan), contains(C,A), on(B,C).}
\end{center}

\noindent This rule expresses the fact, that the cyan square can not be the foundation.
When setting the selection size $k=4$, we let an additional background super-pixel be part of $S$. The resulting rule is:

\begin{center}
	\texttt{concept(A) :- contains(B,A), has\_color(B,cyan), contains(C,A), has\_color(C,blue), top\_of(B,C).}
\end{center}

\noindent This rule captures the fact, that a cyan block has to be above a blue one. The generated explanations are only partial representations of the intended concept. The symbolic explanations capture relevant aspects, but are too general.


\section{Bringing Together Visual and Symbolic Explanations}\label{fusion}

\begin{figure}[t]
	\centering
	\includegraphics[width=0.4\textwidth]{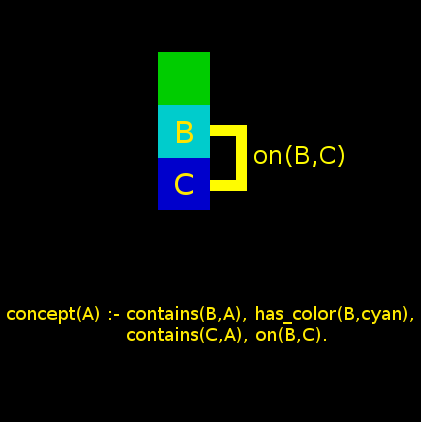}
	\caption{An example for the combination of visual and verbal explanations. Here it is explained, why and where this particular image shows evidence for belonging to the concept tower.} \label{fig:fusion_explanation}
\end{figure}

The generated rules give explanations in symbolic form which can be re-written into verbal statements. We postulate that helpful explanations for images should relate highlighting of relevant parts of the image with explicit symbolic information of attributes and relations. In this section we give an example on how this fusion might look like. Let us take the tower example from Section~\ref{complex_tower}. In Figure~\ref{fig:fusion_explanation}, the output of standard LIME is given with the 3 most important super-pixels matching the expected region in the image. Additionally, the relation from the instantiated rule from the experiment for $k=3$ is given. Since cyan is the only square that is mentioned in the rule, we take it as a reference. The relation \texttt{on} links the cyan square to another unknown square below. This relation is shown explicitly in the image by connecting the two squares and writing the instantiated relation.

\section{Conclusion and Further Work}\label{conclusion}

We proposed an approach to extract symbolic rules from images which can be used to explain classifier decisions in a more expressive way than visual highlighting alone. For a simple artificial domain we gave a proof of concept for our approach LIME-Aleph. The work presented here significantly extends \cite{RaboldSS18} by providing a method of automated extraction of visual attributes and spatial relations from images. As a next step we want to also let the explanative power be evaluated by humans.
Also we plan to cover real world image datasets like explaining differences between towers and houses as shown in Figure~\ref{fig:visualverbal}. The challenge here is to come up with arbitrarily placeable segmentations that are easily interchangeable. While our algorithm relies on a regular grid, in an image ``in the wild'', the semantic borders of sub-objects can be irregular in shape and not easily be flipped in order to test for different relations. One idea to cope with this problems is to use relevance information from inner layers in a CNN (e.g., with LRP,~\cite{samek2017explainable}) to first pinpoint small important regions and sub-objects, then super-imposing a standardized selection shape (square, circle, etc.) over the pixel values to find interchangeable super-pixels for filling selection $S$.

In general, it might be useful to consider a variety of explanation formats to accommodate specific personal preferences and situational contexts. For example, visual highlighting is a quick way to communicate what is important while verbal explanations convey more details. Likewise, examples which prototypically represent a class and near-miss counter-examples could be used to make system decisions more transparent~\cite{adadi2018peeking}.
Explanations might also not be a one-way street. In many domains, it is an illusion that the labeling of the training  is really a ground truth. For example, in medical diagnosis, there are many cases where not even experts agree. Therefore, for many practical applications, learning should be interactive \cite{fails2003interactive}. To constrain model adaption, the user could mark-up that parts of an explanation which are irrelevant or wrong. Such a cooperative approach might improve the joint performance of the human-machine-partnership.

\bibliographystyle{splncs04}
\bibliography{comprehensibleLit}

\end{document}